\documentclass{article}

\usepackage[preprint,nonatbib]{neurips_2023}

\usepackage[utf8]{inputenc} % allow utf-8 input
\usepackage[T1]{fontenc}    % use 8-bit T1 fonts
\usepackage{url}            % simple URL typesetting
\usepackage{booktabs}       % professional-quality tables
\usepackage{amsfonts}       % blackboard math symbols
\usepackage{nicefrac}       % compact symbols for 1/2, etc.
\usepackage{microtype}      % microtypography
\usepackage{xcolor}         % colors
\usepackage{bbding}

\usepackage{color,xcolor}
\usepackage{epsfig}
\usepackage{graphicx}

\usepackage{adjustbox}
\usepackage{array}
\usepackage{booktabs}
\usepackage{colortbl}
\usepackage{float,wrapfig}
\usepackage{hhline}
\usepackage{multirow}
\usepackage{tabularx}
\usepackage{makecell}
\usepackage{float}

\usepackage{amsmath,amsfonts,amsthm,amssymb}
\usepackage{bm}
\usepackage{nicefrac}
\usepackage{microtype}

\usepackage{changepage}
\usepackage{extramarks}
\usepackage{fancyhdr}
\usepackage{lastpage}
\usepackage{setspace}
\usepackage{soul}
\usepackage{xspace}

\usepackage[pagebackref=true,breaklinks=true,colorlinks,citecolor={green},urlcolor={purple}]{hyperref}
\usepackage[nocompress]{cite}
\usepackage{url}

\usepackage{algorithm, algorithmic}
\usepackage{enumerate}
\usepackage{enumitem}
\usepackage{todonotes}

\usepackage{etoolbox,siunitx}
\usepackage{afterpage}
\usepackage{subcaption}
\usepackage[title]{appendix}
\usepackage{overpic}

\usepackage{pifont}
\usepackage{wrapfig}

\newcommand{\ignorethis}[1]{}

\makeatletter
\DeclareRobustCommand\onedot{\futurelet\@let@token\@onedot}
\def\@onedot{\ifx\@let@token.\else.\null\fi\xspace}

\def\eg{\emph{e.g}\onedot} 
\def\ie{\emph{i.e}\onedot} 
 
\def\etc{\emph{etc}\onedot}

\makeatother

\newrobustcmd{\B}{\bfseries}
\newcommand*{\rom}[1]{\expandafter\romannumeral #1}

\definecolor{mydarkblue}{rgb}{0,0.08,1}
\definecolor{mydarkgreen}{rgb}{0.02,0.6,0.02}
\definecolor{mydarkred}{rgb}{0.8,0.02,0.02}
\definecolor{mydarkorange}{rgb}{0.40,0.2,0.02}
\definecolor{mypurple}{RGB}{111,0,255}
\definecolor{myred}{rgb}{1.0,0.0,0.0}
\definecolor{mygold}{rgb}{0.75,0.6,0.12}
\definecolor{myblue}{rgb}{0,0.2,0.8}
\definecolor{mydarkgray}{rgb}{0.66,0.66,0.66}

\newcommand{\myparagraph}[1]{\vspace{-6pt}\paragraph{#1}}

\usepackage{amsmath,amsfonts,bm}

\def\eqref#1{equation~\ref{#1}}
\def\1{\bm{1}}

\def\vc{{\bm{c}}}

\def\vx{{\bm{x}}}

\def\vz{{\bm{z}}}

\def\mK{{\bm{K}}}

\def\mQ{{\bm{Q}}}

\def\mV{{\bm{V}}}
\def\mW{{\bm{W}}}

\DeclareMathAlphabet{\mathsfit}{\encodingdefault}{\sfdefault}{m}{sl}
\SetMathAlphabet{\mathsfit}{bold}{\encodingdefault}{\sfdefault}{bx}{n}

\newcommand{\E}{\mathbb{E}}
\newcommand{\Ls}{\mathcal{L}}

\newcommand{\encoder}{\mathcal{E}}
\newcommand{\decoder}{\mathcal{D}}

\title{ControlVideo: Training-free Controllable Text-to-Video Generation}

\author{%
Yabo Zhang$^1$ \ Yuxiang Wei$^1$ \ Dongsheng Jiang$^2$ \ Xiaopeng Zhang$^2$ \ Wangmeng Zuo$^{1}$ $^{(}$\Envelope$^)$ Qi Tian$^2$ 
\\
\\
\textsuperscript{1}Harbin Institute of Technology
\textsuperscript{2}Huawei Cloud
}

\begin{document}

\maketitle

\begin{abstract}
Text-driven diffusion models have unlocked unprecedented abilities in image generation, whereas their video counterpart still lags behind due to the excessive training cost of temporal modeling. Besides the training burden, the generated videos also suffer from appearance inconsistency and structural flickers, especially in long video synthesis.
To address these challenges, we design a \emph{training-free} framework called \textbf{ControlVideo} to enable natural and efficient text-to-video generation.    
ControlVideo, adapted from ControlNet, leverages coarsely structural consistency from input motion sequences, and introduces three modules to improve video generation. 
Firstly, to ensure appearance coherence between frames, ControlVideo adds fully cross-frame interaction in self-attention modules. 
Secondly, to mitigate the flicker effect, it introduces an interleaved-frame smoother that employs frame interpolation on alternated frames. 
Finally, to produce long videos efficiently, it utilizes a hierarchical sampler that separately synthesizes each short clip with holistic coherency.
Empowered with these modules, ControlVideo
outperforms the state-of-the-arts on extensive motion-prompt pairs quantitatively and qualitatively.
Notably, thanks to the efficient designs, it generates both short and long videos within several minutes using one NVIDIA 2080Ti.
% Code will be publicly available.
Code is available at \url{https://github.com/YBYBZhang/ControlVideo}.
\end{abstract}

\section{Introduction}
\begin{figure}[t]
   \begin{center}
   \includegraphics[width=.99\linewidth]{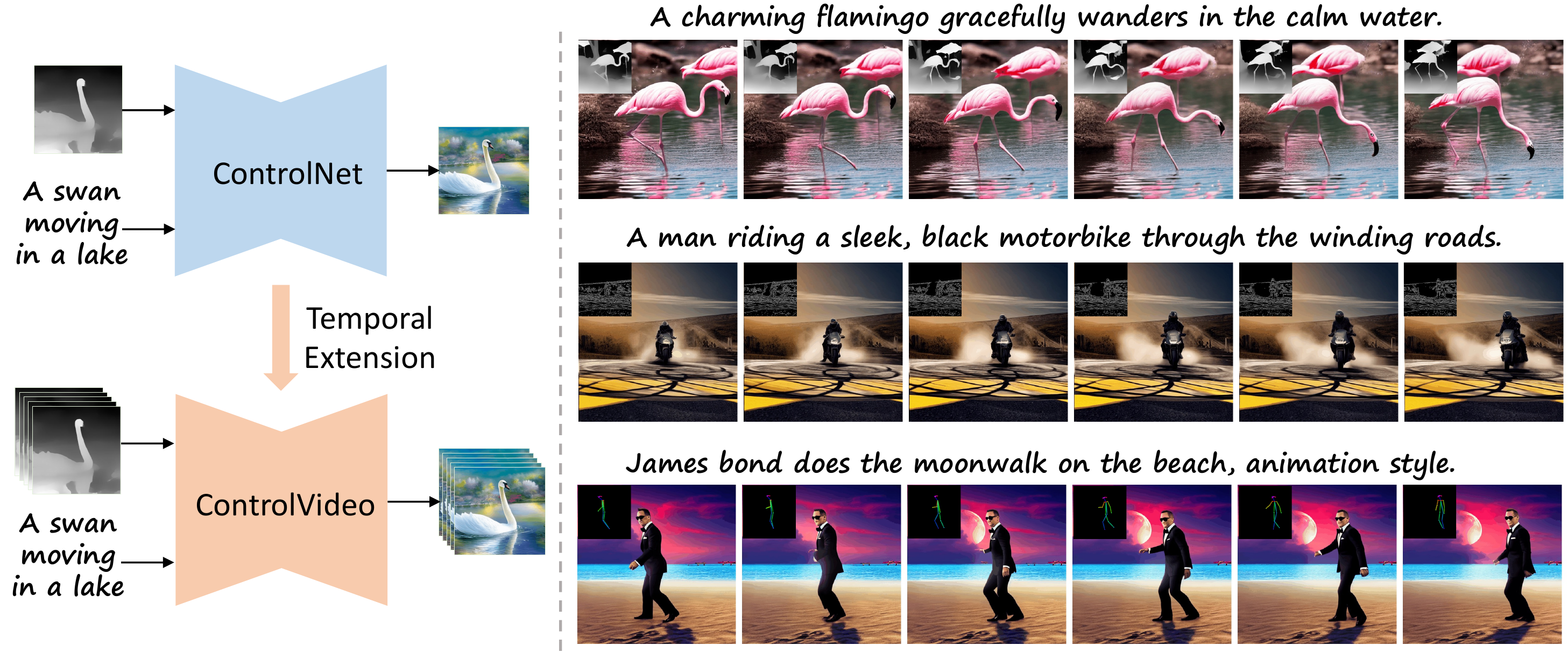}
   \end{center}
  \vspace{-0.5em}
   \caption{\textbf{Training-free controllable text-to-video generation.} 
   \textbf{Left: }ControlVideo adapts ControlNet to the video counterpart by inflating along the temporal axis, aiming to directly inherit its high-quality and consistent generation without any finetuning.
   \textbf{Right: }ControlVideo could synthesize photo-realistic videos conditioned on various motion sequences, which are temporally consistent in both structure and appearance.
   \textbf{Results best seen at 500\% zoom.}} 
    \label{fig:intro}
    \vspace{-1.5em}
\end{figure}

Large-scale diffusion models have made a tremendous breakthrough on text-to-image synthesis~\cite{nichol2021glide,rombach2022high,balaji2022ediffi,ramesh2022hierarchical,saharia2022photorealistic} and their creative applications~\cite{gal2022image,wei2023elite,ni2022imaginarynet,hertz2022prompt}.
Several works~\cite{video-diffusion-models,imagen_video,make-a-video,gen1_paper,hong2022cogvideo} attempt to replicate this success in the video counterpart, \ie, modeling higher-dimensional complex video distributions in the wild world.
However, training such a text-to-video model requires massive amounts of high-quality videos and computational resources, 
which limits the further research and applications by relevant communities.

% To reduce the excessively training requirements, we study a new and efficient form: \textit{controllable text-to-video generation}.
%
To reduce the excessive training requirements, we study a new and efficient form: \textit{controllable text-to-video generation with text-to-image models}.
This task aims to produce a video conditioned on both a textual description and motion sequences (\eg, depth or edge maps).
%
% Instead of learning the video distribution from scratch, it could efficiently leverage prior knowledge from pre-trained text-to-image generative models~\cite{rombach2022high,ramesh2022hierarchical} and motion sequences, 
% \ie, the former enable the capability of text-driven creation while the latter provide coarsely temporal consistency in structure.
%
As shown in Fig.~\ref{fig:intro}, instead of learning the video distribution from scratch, it could efficiently leverage the generation capability of pre-trained text-to-image generative models~\cite{rombach2022high,ramesh2022hierarchical} and coarsely temporal consistency of motion sequences to produce vivid videos.

% Recent studies~\cite{tune-a-video,khachatryan2023text2video,ma2023follow} explore leveraging the pre-trained text-to-image model for video generation. 
% With the structure-controllable power of ControlNet~\cite{controlnet}, researchers and users leverage sequential motion information (\eg, depth or edge maps) to synthesize videos with coarse temporal consistency.
% This process requires little additional finetuning and thus significantly reduces heavy training burden of text-to-video models.
% We study this new and efficient form: \textit{controllable text-to-video generation}.

Recent studies~\cite{tune-a-video,khachatryan2023text2video} have explored leveraging the structure controllability of ControlNet~\cite{controlnet} or DDIM inversion~\cite{DDIM_paper} for video generation.
Rather than synthesizing all frames independently,
~\cite{tune-a-video,khachatryan2023text2video} enhance appearance coherence by replacing original self-attention with the sparser cross-frame attention. 
Nevertheless, their video quality is still far behind photo-realistic videos in terms of: (i) inconsistent appearance between some frames (see Fig.~\ref{fig:main_depth_canny} (a)),
(ii) visible artifacts in large motion videos (see Fig.~\ref{fig:main_depth_canny} (b)),
\textit{and} (iii) structural flickers during inter-frame transitions.
For (i) and (ii), their sparser cross-frame mechanisms increase the discrepancy between the query and key in self-attention modules, and hence impede inheriting high-quality and consistent generation from pre-trained text-to-image models.
For (iii), input motion sequences only provide the coarse-level structure of videos, failing to smoothly transition between consecutive frames.
% Besides, when existing methods synthesize longer videos, 
% the length of synthesized videos becomes longer, both the GPU memory requirement and temporal inconsistency also increase greatly, which further impede in practical applications.

In this work, we propose a training-free \textit{ControlVideo} for high-quality and consistent controllable text-to-video generation, along with \textit{interleaved-frame smoother} to enhance structural smoothness. 
\textit{ControlVideo} directly inherits the architecture and weights from ControlNet~\cite{controlnet}, while adapting it to the video counterpart by extending self-attention with the \textit{fully cross-frame interaction}.
Different from prior works~\cite{tune-a-video,khachatryan2023text2video}, our fully cross-frame interaction concatenates all frames to become a ``larger image'', thus directly inheriting high-quality and consistent generation from ControlNet.
\textit{Interleaved-frame smoother} deflickers the whole video via the interleaved interpolation at selected sequential timesteps. 
%
% As illustrated in Fig.~\ref{fig:smoother},
% sequential operations alternately interpolate even or odd frames to smooth the interleaved three-frame clips, and their combination makes the entire video stable.
As illustrated in Fig.~\ref{fig:smoother}, the operation at each timestep smooths the interleaved three-frame clips by interpolating middle frames, and the combination at two consecutive timesteps smooths the entire video.
% \tocheck{
% As illustrated in Fig.~\ref{fig:smoother},
% each operation smooths the interleaved three-frame clips through interpolation.
% %
% By alternately interpolating even or odd frames at different timesteps, it can smooth the entire video and generate visually consistent results.
% }
%
% Notably, the smoothing operation is performed on the predicted RGB frames at selected timesteps, 
Since the smoothing operation is only performed at a few timesteps, the quality and individuality of interpolated frames can be well retained by the following denoising steps.
% thus its side effect (\eg, blur and distortion) can be effectively reduced by the following denoising steps.

To enable efficient long-video synthesis, we further introduce a \textit{hierarchical sampler} to produce separated short clips with long-term coherency.
In specific, a long video is first split into multiple short video clips with the selected key frames.
Then, the key frames are pre-generated with fully cross-frame attention for long-range coherence.
Conditioned on pairs of key frames, we sequentially synthesize their corresponding intermediate short video clips with the global consistency.
% Finally, we synthesize each short video in a clip-by-clip way, where the interval frames in each clip are conditioned on their adjacent key frames only.

We conduct the experiments on extensively collected motion-prompt pairs.
The experimental results show that our method outperforms alternative competitors qualitatively and quantitatively.
Thanks to the efficient designs, \ie, the xFormers~\cite{xformers} implementation and hierarchical sampler, ControlVideo can produce both short and long videos within several minutes using one NVIDIA $2080$Ti.

In summary, our contributions are presented as follows:

\vspace{-3mm}
% \begin{itemize}[itemsep=0pt, parsep=0pt, leftmargin=10pt]
%     \item We introduce training-free ControlVideo for controllable text-to-video generation, and our fully cross-attention demonstrates better performance in quality and consistency.
%     \item We propose interleaved-frame smoother to reduce structural flickers throughout a whole video.
%     \item We present the hierarchical sampling to enable efficient long-video generation in commodity GPUs.
% \end{itemize}
% \tocheck{
\begin{itemize}[itemsep=0pt, parsep=0pt, leftmargin=10pt]
    \item We propose a training-free ControlVideo for controllable text-to-video generation, which consists of the fully cross-frame interaction, interleaved-frame smoother, and hierarchical sampler.
    \item The fully cross-attention demonstrates higher video quality and appearance consistency, while interleaved-frame smoother further reduces structural flickers throughout a whole video.
    \item The hierarchical sampler enables efficient long-video generation in commodity GPUs.
\end{itemize}
% }
\section{Background}
% 1. latent diffusion
%     1.1 contain self-attn (opt)
%     1.2 ddim and xt -> x0
% 2. ControlNet
% \textbf{Diffusion Models} (DMs) learn to model the data distribution by 
% LDM transforms a data distribution into a normal gaussian distribution with a forward pass, while generates learns to reverse this process with a model $\epsilon_{\theta}$.
% LDM slowly adds the gaussian noise 
% LDM performs both forward and back

\textbf{Latent diffusion model} (LDM)~\cite{rombach2022high} is an efficient variant of diffusion models~\cite{DDPM_paper} by applying the diffusion process in the latent space rather than image space.
LDM contains two main components.
Firstly, it uses an encoder $\encoder$ to compress an image $\vx$ into latent code $\vz=\encoder(\vx)$ and a decoder to reconstruct this image $\vx \approx \decoder(\vz)$, respectively.
Secondly, it learns the distribution of image latent codes $\vz_0 \sim p_{data}(\vz_0)$ in a DDPM formulation~\cite{DDPM_paper}, including a forward and a backward process.
The forward diffusion process gradually adds gaussian noise at each timestep $t$ to obtain $\vz_t$:
\begin{equation}
    q(\vz_t|\vz_{t-1}) = \mathcal{N}(\vz_t; \sqrt{1-\beta_t}\vz_{t-1}, \beta_t I),
    \label{eq:forward}
\end{equation}
where $\{\beta_{t}\}_{t=1}^{T}$ are the scale of noises, and $T$ denotes the number of diffusion timesteps.
The backward denoising process reverses the above diffusion process to predict less noisy $\vz_{t-1}$:
\begin{equation}
    p_\theta(\vz_{t-1}|\vz_t) = \mathcal{N}(\vz_{t-1};\mu_\theta(\vz_t,t),\Sigma_\theta(\vz_t,t)).
    \label{eq:backward}
\end{equation}
The $\mu_\theta$ and $\Sigma_\theta$ are implemented with a denoising model $\epsilon_{\theta}$ with learnable parameters $\theta$, which is trained with a simple objective:
\begin{align}
    \Ls_{simple} := \E_{\encoder(\vz), \epsilon \sim \mathcal{N}(0, 1),  t}\Big[ \Vert \epsilon - \epsilon_{\theta}(\vz_{t},t) \Vert_{2}^{2}\Big].
    \label{eq:ldmloss}
\end{align}
When generating new samples, we start from $\vz_{T} \sim \mathcal{N}(0, 1)$ and employ DDIM sampling to predict $\vz_{t-1}$ of previous timestep:
\begin{align}
    \vz_{t-1} & = \sqrt{\alpha_{t-1}} \underbrace{\left(\frac{\vz_t - \sqrt{1 - \alpha_t} \epsilon_\theta(\vz_t, t)}{\sqrt{\alpha_t}}\right)}_{\text{`` predicted } \vz_{0} \text{''}} + \underbrace{\sqrt{1 - \alpha_{t-1}} \cdot \epsilon_\theta(\vz_t, t)}_{\text{``direction pointing to } \vz_t \text{''}},
    \label{eq:ddim}
\end{align}
where $\alpha_{t} = \prod_{i=1}^{t}(1-\beta_i)$. 
We use $\vz_{t \rightarrow 0}$ to represent ``predicted $\vz_{0}$'' at timestep $t$ for simplicity.
Note that we use Stable Diffusion (SD) $\epsilon_\theta(\vz_t, t, \tau)$ as our base model, which is an instantiation of text-guided LDMs pre-trained on billions of image-text pairs.
$\tau$ denotes the text prompt.

\textbf{ControlNet}~\cite{controlnet} enables SD to support more controllable input conditions during text-to-image synthesis, \eg, depth maps, poses, edges, \etc.
The ControlNet uses the same U-Net~\cite{UNet_paper} architecture as SD and finetunes its weights to support task-specific conditions, converting $\epsilon_\theta(\vz_t, t, \tau)$ to $\epsilon_\theta(\vz_t, t, \vc, \tau)$, where $\vc$ denotes additional conditions.
To distinguish the U-Net architectures of SD and ControlNet, we denote the former as the \textit{main U-Net} while the latter as the \textit{auxiliary U-Net}.

\section{ControlVideo}
\label{sec:method}
\begin{figure}[t]
   % \vspace{-1em}
   \begin{center}
   \includegraphics[width=.95\linewidth]{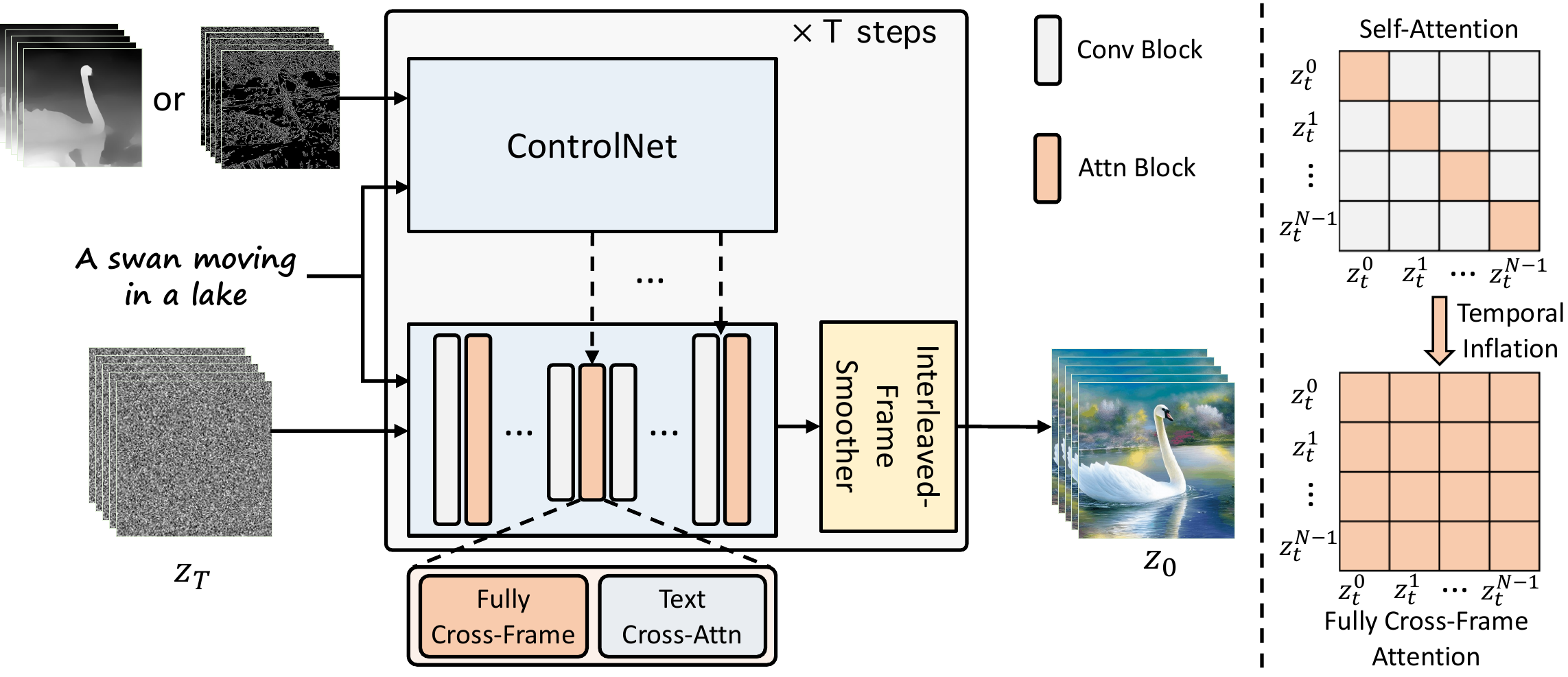}
   % \fbox{\rule{0pt}{2in} \rule{.9\linewidth}{0pt}} 
   \end{center}
   \vspace{-2mm}
   \caption{
   \textbf{Overview of ControlVideo.}
   For consistency in appearance, ControlVideo adapts ControlNet to the video counterpart by adding fully cross-frame interaction into self-attention modules.
   Considering the flickers in structure, the interleaved-frame smoother is integrated to smooth all inter-frame transitions via the interleaved interpolation (see Fig.~\ref{fig:smoother} for details).
   }
    \label{fig:controlvideo}
    % \vspace{-4mm}
\end{figure}

Controllable text-to-video generation aims to produce a video of length $N$ conditioned on motion sequences $\vc=\{\vc^i\}_{i=0}^{N-1}$ and a text prompt $\tau$.
As illustrated in Fig.~\ref{fig:controlvideo}, we propose a training-free framework termed ControlVideo towards consistent and efficient video generation.
% which is adapted from the image counterpart ControlNet.
Firstly, ControlVideo is adapted from ControlNet by employing \textit{fully cross-frame interaction}, which ensures the appearance consistency with less quality degradation.
% without sacrificing high-quality generation.
%
% Moreover, we present the interleaved-frame smoother to smooth the interleaved three-frame clips, thereby reducing the structural flickers throughout the whole video.
%这句和上面的二选一
Secondly, the \textit{interleaved-frame smoother} deflickers the whole video by interpolating alternate frames at sequential timesteps.
% To reduce the structural flickers across the whole video,
% we further present an interleaved-frame smoother module to smooth the interleaved three-frame clips.
%
Finally, the \textit{hierarchical sampler} separately produces short clips with the holistic coherency to enable long video synthesis.
% is adopted to efficiently produce a long video clip-by-clip. 
% Each module would be elaborated as follows.

\vspace{0.5em}
\myparagraph{Fully cross-frame interaction.} 
% \textcolor{red}{use another name like "fully cross-frame attention " }
The main challenge of adapting text-to-image models to the video counterpart is to ensure temporal consistency.
Leveraging the controllability of ControlNet, motion sequences could provide coarse-level consistency in structure.
Nonetheless, even using the same initial noise, individually producing all frames with ControlNet will lead to drastic inconsistency in appearance (see row $2$ in Fig.~\ref{fig:ab_crossframe}).
To keep the video appearance coherent, we concatenate all video frames to become a ``large image'', so that their content could be shared via inter-frame interaction.
Considering that self-attention in SD is driven by appearance similarities~\cite{tune-a-video}, we propose to enhance the holistic coherency by adding attention-based fully cross-frame interaction.
% ControlVideo is extended from the image counterpart ControlNet by inflating the main U-Net along the temporal axis.

In specific, ControlVideo inflates the main U-Net from Stable Diffusion along the temporal axis, while keeping the auxiliary U-Net from ControlNet.
Analogous to~\cite{video-diffusion-models,tune-a-video,khachatryan2023text2video}, it directly converts 2D convolution layers to 3D counterpart by replacing $3 \times 3$ kernels with $1\times 3 \times 3$ kernels.
In Fig.~\ref{fig:controlvideo} (right), it extends self-attention by adding interaction across all frames:
{\small
\begin{equation}
    \mathrm{Attention}(\mQ,\mK,\mV)=\mathrm{Softmax}(\frac{\mQ \mK^T}{\sqrt{d}}) \cdot \mV,
    \text{ where }\mQ=\mW^Q \vz_t, \  \mK=\mW^K \vz_t,\  \mV=\mW^V \vz_t,
    \label{eq:full_attn}
\end{equation}
}
where $\vz_t = \{\vz_t^i\}_{i=0}^{N-1}$ denotes all latent frames at timestep $t$, while $\mW^Q$, $\mW^K$, and $\mW^V$ project $\vz_t$ into query, key, and value, respectively.

Previous works~\cite{tune-a-video,khachatryan2023text2video} usually replace self-attention with sparser cross-frame mechanisms, \eg, all frames attend to the first frame only.
Yet, these mechanisms will increase the discrepancy between the query and key in self-attention modules, resulting in the degradation of video quality and consistency.
%
% especially in large motion videos.
In comparison, our fully cross-frame mechanism combines all frames into a ``large image'', and has a less generation gap with text-to-image models (see comparisons in Fig.~\ref{fig:ab_crossframe}).
Moreover, with the efficient implementation, the fully cross-frame attention only brings little memory and acceptable computational burden in short-video generation ($< 16$ frames).

\myparagraph{Interleaved-frame smoother.}
\begin{wrapfigure}[22]{o}{0.55\textwidth}
  % \vspace{-.3cm}
  \begin{center}
   \includegraphics[width=.99\linewidth]{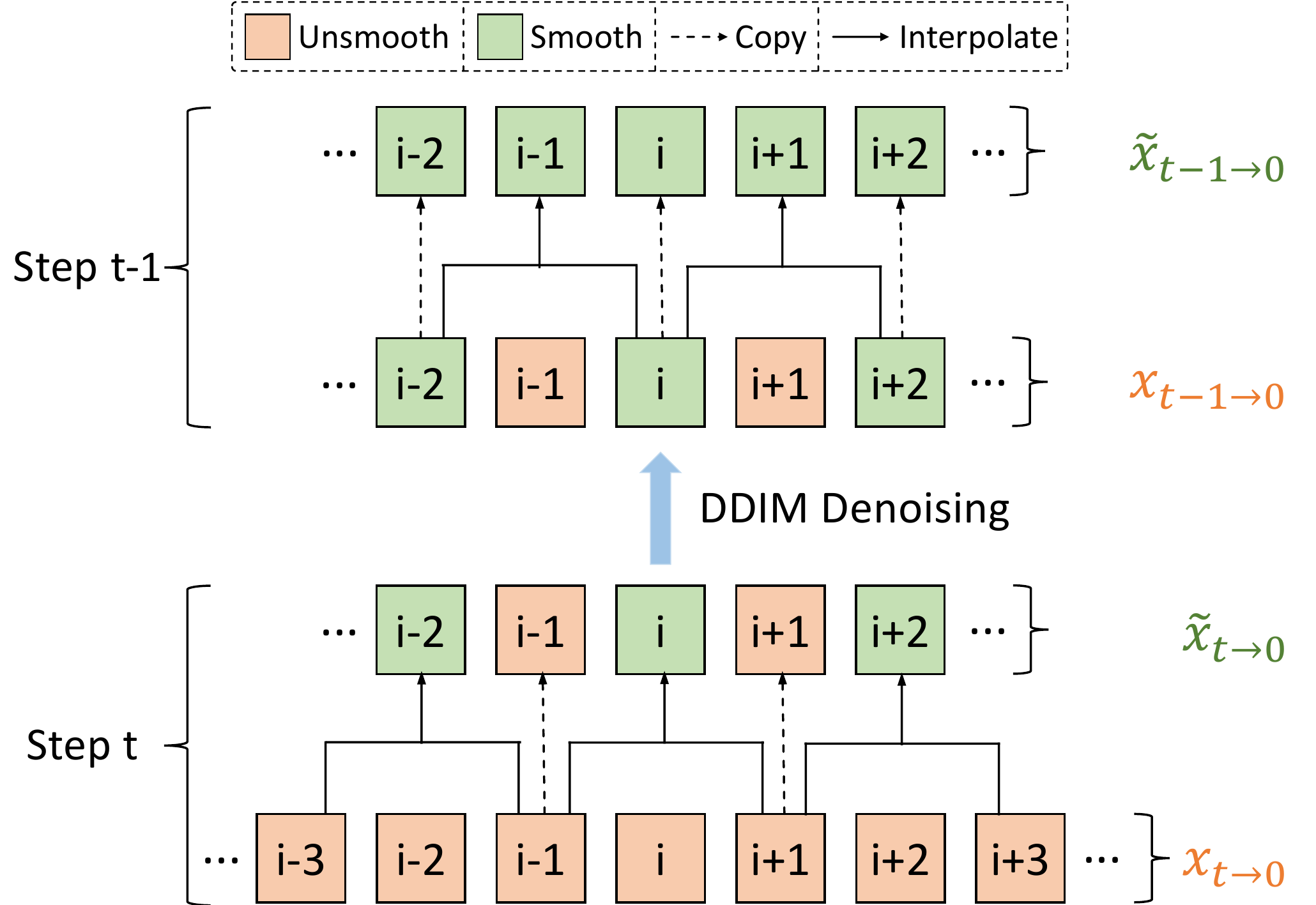}
   \end{center}
   % \vspace{-2mm}
   \caption{
   \textbf{Illustration of interleaved-frame smoother.}
   At timestep $t$, predicted RGB frames \textcolor{orange}{$\vx_{t\rightarrow 0}$} are smoothed into \textcolor[RGB]{0, 100, 0}{$\tilde{\vx}_{t\rightarrow 0}$} via middle-frame interpolation.
   The combination of two sequential timesteps reduces the structural flickers over the entire video.
   }
    \label{fig:smoother}
    % \vspace{-2mm}
\end{wrapfigure}

Albeit the videos produced by fully cross-frame interaction are promisingly consistent in appearance, they are still visibly flickering in structure.
Input motion sequences only ensure the synthesized videos with coarse-level structural consistency, but not enough to keep the smooth transition between consecutive frames.
Therefore, we further propose an interleaved-frame smoother to mitigate the flicker effect in structure.
As shown in Fig.~\ref{fig:smoother}, our key idea is to smooth each three-frame clip by interpolating the middle frame, following by repeating it in an interleaved manner to smooth the whole video.

% \tocheck{Specifically, the interleaved-frame smoother alternately interpolates the even or odd frames at sequential timesteps, so that smoothing their corresponding three-frame clips.
% where
% each interpolation smooths an even or odd three-frame clip.
% % , where every smoothed clip consists of the interpolated frame with its adjacent two frames.
% For two sequential timesteps, the combination of their smoothed three-frame clips could be regarded as deflickering the whole video.}

Specifically, our interleaved-frame smoother is performed on predicted RGB frames at sequential timesteps.
The operation at each timestep interpolates the even or odd frames to smooth their corresponding three-frame clips.
In this way, the smoothed three-frame clips from two consecutive timesteps are overlapped together to deflicker the entire video.
Before applying our interleaved-frame smoother at timestep $t$, we first predict the clean video latent $\vz_{t\rightarrow 0}$ according to $\vz_t$:
\begin{align}
    \vz_{t\rightarrow 0} = \frac{\vz_t - \sqrt{1 - \alpha_t} \epsilon_\theta(\vz_t, t, \vc, \tau)}{\sqrt{\alpha_t}}.
    \label{eq:pred_clean}
\end{align}

After projecting $\vz_{t\rightarrow 0}$ into a RGB video $\vx_{t\rightarrow 0} = \mathcal{D}(\vz_{t\rightarrow 0})$, we convert it to a more smoothed video $\tilde{\vx}_{t\rightarrow 0}$ using our interleaved-frame smoother.
Based on smoothed video latent $\tilde{\vz}_{t\rightarrow 0} = \mathcal{E}(\tilde{\vx}_{t\rightarrow 0})$, we compute the less noisy latent $\vz_{t-1}$ following DDIM denoising in Eq.~\ref{eq:ddim}:
\begin{align}
    \vz_{t-1} & = \sqrt{\alpha_{t-1}} \tilde{\vz}_{t\rightarrow 0} + \sqrt{1 - \alpha_{t-1}} \cdot \epsilon_\theta(\vz_t, t, \vc, \tau).
    \label{eq:smooth_ddim}
\end{align}

Notably, the above process is only performed at the selected intermediate timesteps, which has two advantages: (i) the newly computational burden can be negligible 
\textit{and} (ii) the individuality and quality of interpolated frames are well retained by the following denoising steps.

\myparagraph{Hierarchical sampler.}
Since video diffusion models need to maintain the temporal consistency with inter-frame interaction, they often require substantial GPU memory and computational resources, especially when producing longer videos.
To facilitate efficient and consistent long-video synthesis, we introduce a hierarchical sampler to produce long videos in a clip-by-clip manner.
At each timestep, a long video $\vz_t = \{\vz_t^i\}_{i=0}^{N-1}$ is separated into multiple short video clips
% $\{\vz_t^{k N_c : (k+1) N_c}\}_{k=0}^{\frac {N} {N_c} - 1}$ 
with the selected key frames $\vz_t^{key}=\{\vz_t^{k N_c}\}_{k=0}^{\frac {N} {N_c}}$, where each clip is of length $N_c - 1$ and the $k$th clip is denoted as $\widehat{\vz}_t^{k}=\{\vz_t^j\}_{j=k N_c +1} ^ {(k+1) N_c -1}$.
Then, we pre-generate the key frames with \textit{fully} cross-frame attention for long-range coherence, and their query, key, and value are computed as:
\begin{equation}
    \mQ^{key}=\mW^Q \vz_t^{key},\ \mK^{key}=\mW^K \vz_t^{key},\ \mV^{key}=\mW^V \vz_t^{key}.
    \label{eq:key_frame}
\end{equation}
Conditioned on each pair of key frames, we sequentially synthesize their corresponding clips holding the holistic consistency:
\begin{equation}
    \widehat{\mQ}^{k}=\mW^Q \widehat{\vz}_t^{k}, \quad \widehat{\mK}^{k}=\mW^K [\vz_t^{k N_c}, \vz_t^{(k+1) N_c}], \quad \widehat{\mV}^{k}=\mW^V [\vz_t^{k N_c}, \vz_t^{(k+1) N_c}].
    \label{eq:clip}
\end{equation}

\section{Experiments}
\begin{figure}[t]
%   \vspace{-1em}
  \begin{center}
  \includegraphics[width=.99\linewidth]{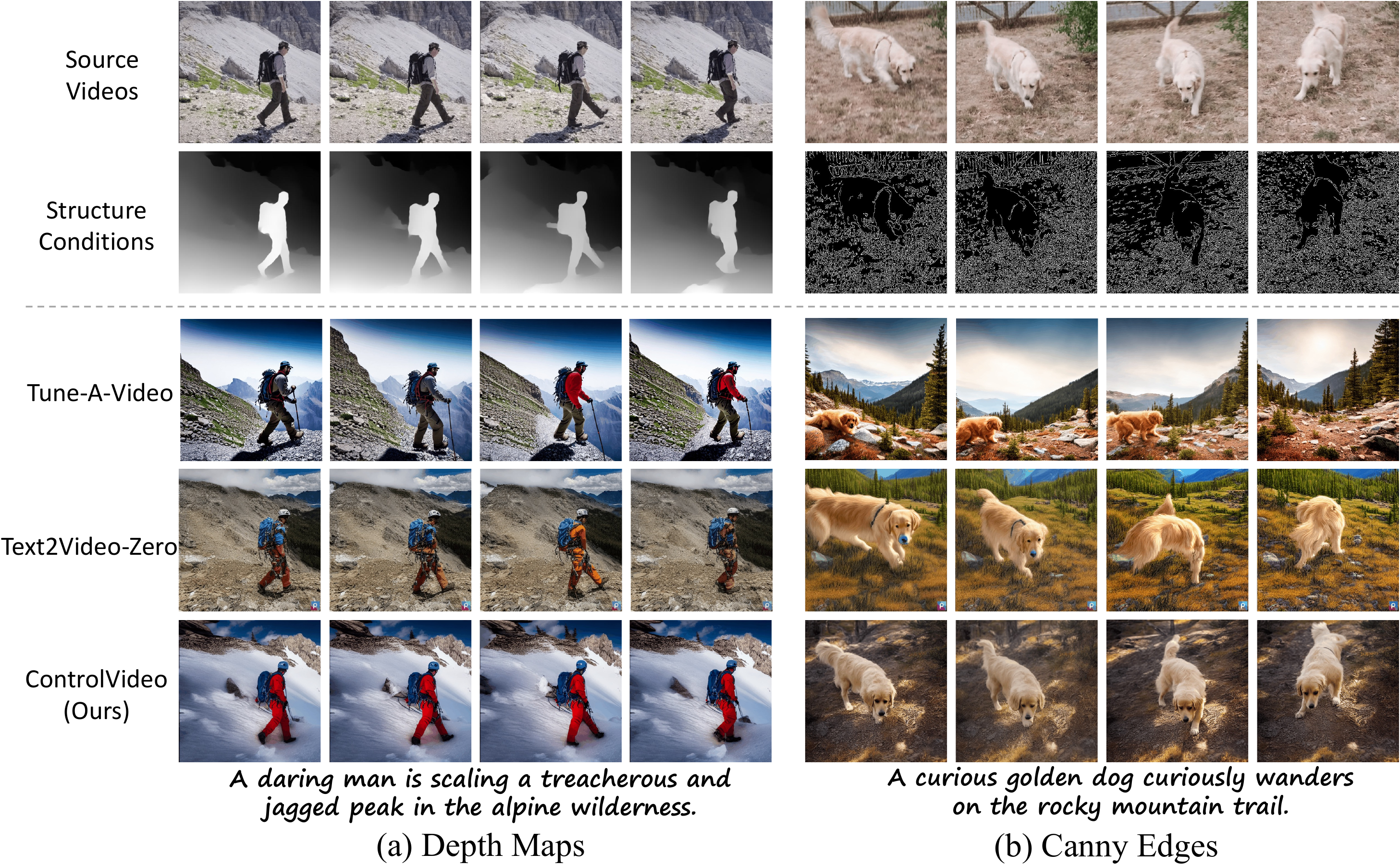}
  \end{center}
  \vspace{-1mm}
  \caption{
  \textbf{Qualitative comparisons conditioned on depth maps and canny edges.}
  Our ControlVideo produces videos with better (a) appearance consistency and (b) video quality than others.
  In contrast, Tune-A-Video~\cite{tune-a-video} fails to inherit structures from source videos, while Text2Video-Zero~\cite{khachatryan2023text2video} brings visible artifacts in large motion videos.
  \textbf{Results best seen at 500\% zoom.}
  }
    \label{fig:main_depth_canny}
  \vspace{-1mm}
\end{figure}
\begin{table}[t]
% \vspace{-1em}
\caption{
\textbf{Quantitative comparisons} of ControlVideo  with other methods.
We evaluate them on $125$ motion-prompt pairs in terms of consistency, and the best results are \textbf{bolded}.
}
\vspace{2mm}
\centering
\scalebox{0.9}{
\begin{tabular}{lccc}
\toprule
Method & Structure Condition & Frame Consistency ($\%$) & Prompt Consistency ($\%$)
\\
\midrule
Tune-A-Video~\cite{tune-a-video} & DDIM Inversion~\cite{DDIM_paper} &{94.53} &{31.57} \\
\midrule
Text2Video-Zero~\cite{khachatryan2023text2video} & Canny Edge &{95.17} &{30.74} \\
\textbf{ControlVideo} &Canny Edge &\textbf{96.83} &\textbf{30.75} \\
\midrule
Text2Video-Zero~\cite{khachatryan2023text2video} & Depth Map &{95.99} &{31.69} \\
\textbf{ControlVideo} &Depth Map &\textbf{97.22} &\textbf{31.81} \\
\bottomrule
\end{tabular}
}
\label{tab:main_consistency}
% \vspace{-4mm}
\end{table}

\subsection{Experimental Settings}

\myparagraph{Implementation details.}
Our ControlVideo is adapted from
ControlNet
\footnote{https://huggingface.co/lllyasviel/ControlNet}~\cite{controlnet}
, and our interleaved-frame smoother employs a lightweight RIFE~\cite{huang2022rife} to interpolate the middle frame of each three-frame clip.
The synthesized short videos are of length $15$, while the long videos usually contain about $100$ frames.
Unless otherwise noted, their resolution is both $512 \times 512$. 
During sampling, we adopt DDIM sampling~\cite{DDIM_paper} with $50$ timesteps, and interleaved-frame smoother is performed on predicted RGB frames at timesteps $\{30,31\}$ by default.
%
% Our ControlVideo is efficiently implemented with xFormers~\cite{xformers}.
% %
% Thus, both short and long videos can be produced with one NVIDIA $2080$Ti, where they only take about $2$ and $10$ minutes, respectively.
%
With the efficient implementation of xFormers~\cite{xformers}, our ControVideo could produce both short and long videos with one NVIDIA RTX $2080$Ti in about $2$ and $10$ minutes, respectively.

\myparagraph{Datasets.}
To evaluate our ControlVideo, we collect $25$ object-centric videos from DAVIS dataset~\cite{pont20172017} and manually annotate their source descriptions.
%
% Then, according to these source descriptions, we ask ChatGPT~\cite{openai2022chatgpt} to automatically design five edited prompts for each of them, resulting in $125$ video-prompt pairs in total.
%
Then, for each source description, ChatGPT~\cite{openai2022chatgpt} is utilized to generate five editing prompts automatically, resulting in $125$ video-prompt pairs in total.
Finally, we employ Canny and MiDaS DPT-Hybrid model~\cite{Ranftl2020} to estimate the edges and depth maps of source videos, and form $125$ motion-prompt pairs as our evaluation dataset.
More details are provided in the supplementary materials.

\myparagraph{Metrics.}
Following~\cite{gen1_paper,tune-a-video}, we adopt CLIP~\cite{radford2021learning} to evaluate the video quality from two perspectives.
(i) Frame Consistency: the average cosine similarity between all pairs of consecutive frames, \textit{and}
(ii) Prompt Consistency: the average cosine similarity between input prompt and all video frames.

\myparagraph{Baselines.}
We compare our ControlVideo with three publicly available methods:
(i) Tune-A-Video~\cite{tune-a-video} extends Stable Diffusion to the video counterpart by finetuning it on a source video.
During inference, it uses the DDIM inversion codes of source videos to provide structure guidance.
(ii) Text2Video-Zero~\cite{khachatryan2023text2video} is based on ControlNet, and employs the first-only cross-frame attention on Stable Diffusion without finetuning.
(iii) Follow-Your-Pose~\cite{ma2023follow} is initialized with Stable Diffusion, and is finetuned on LAION-Pose~\cite{ma2023follow} to support human pose conditions.
After that, it is trained on millions of videos~\cite{xue2022advancing} to enable temporally-consistent video generation.
% , it also uses temporal attention and modules like sparse-causal cross-frame attention, but 

\subsection{Qualitative and quantitative comparisons}
\vspace{6pt}
\myparagraph{Qualitative results.}
Fig.~\ref{fig:main_depth_canny} first illustrates the visual comparisons of synthesized videos conditioned on both (a) depth maps and (b) canny edges.
As shown in Fig.~\ref{fig:main_depth_canny} (a), our ControlVideo demonstrates better consistency in both appearance and structure than alternative competitors.
Tune-A-Video fails to keep the temporal consistency of both appearance and fine-grained structure, \eg, \texttt{the color of coat and the structure of road}.
With the motion information from depth maps, Text2Video-Zero achieves promising consistency in structure, but still struggles with incoherent appearance in videos \eg, \texttt{the color of coat}.
%
% In Fig.~\ref{fig:main_depth_canny} (b)
Besides, our ControlVideo also performs more robustly when dealing with large motion inputs.
As illustrated in Fig.~\ref{fig:main_depth_canny} (b), Tune-A-Video ignores the structure information from source videos.
%
% Text2Video-Zero  generates later frames with visible artifacts, since its first-only cross-frame mechanism trades off frame quality for appearance consistency.
%
Text2Video-Zero adopts the first-only cross-frame mechanism to trade off frame quality and appearance consistency, and generates later frames with visible artifacts.
%
% For the fully cross-frame mechanism in ControlVideo, all frames are concatenated together to become a ``larger image'', thereby directly inheriting high quality from text-to-image model.
%
In contrast, with the proposed fully cross-frame mechanism and interleaved-frame smoother, our ControlVideo can handle large motion to generate high-quality and consistent videos.

\begin{figure}[t]
\centering
\begin{minipage}{.48\textwidth}
    \centering
    \includegraphics[width=.99\linewidth]{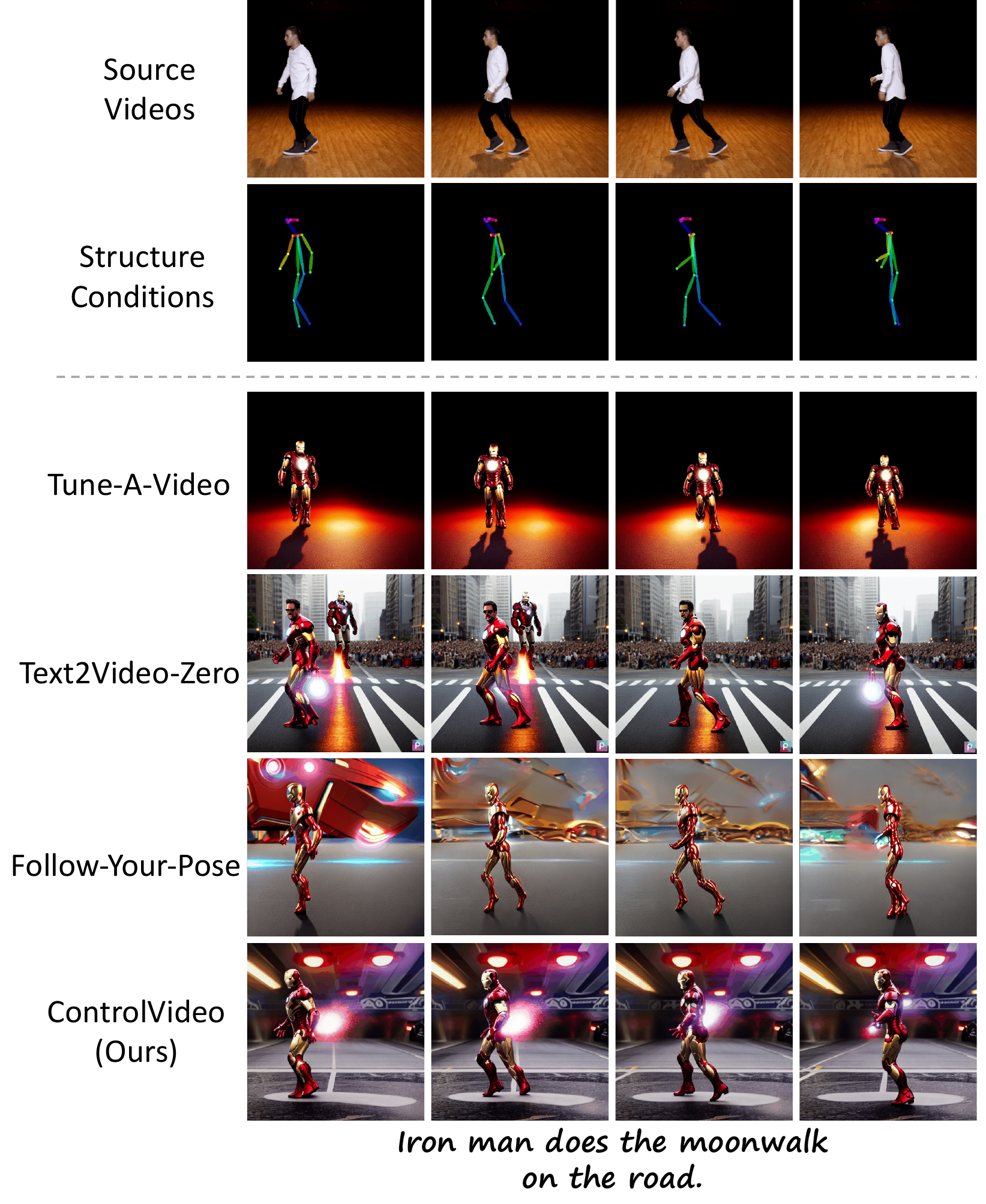}
    \caption{\textbf{Qualitative comparisons on poses}.
    Tune-A-Video~\cite{tune-a-video} only preserves original human positions, while Text2Video-Zero~\cite{khachatryan2023text2video} and Follow-Your-Pose~\cite{ma2023follow} produce frames with appearance incoherence, \eg, changing faces of iron man.
    Our ControlVideo achieves better consistency in both structure and appearance.}
    \label{fig:main_pose}
\end{minipage}
\hfill
\begin{minipage}{.48\textwidth}
    \centering
    \includegraphics[width=.93\linewidth]{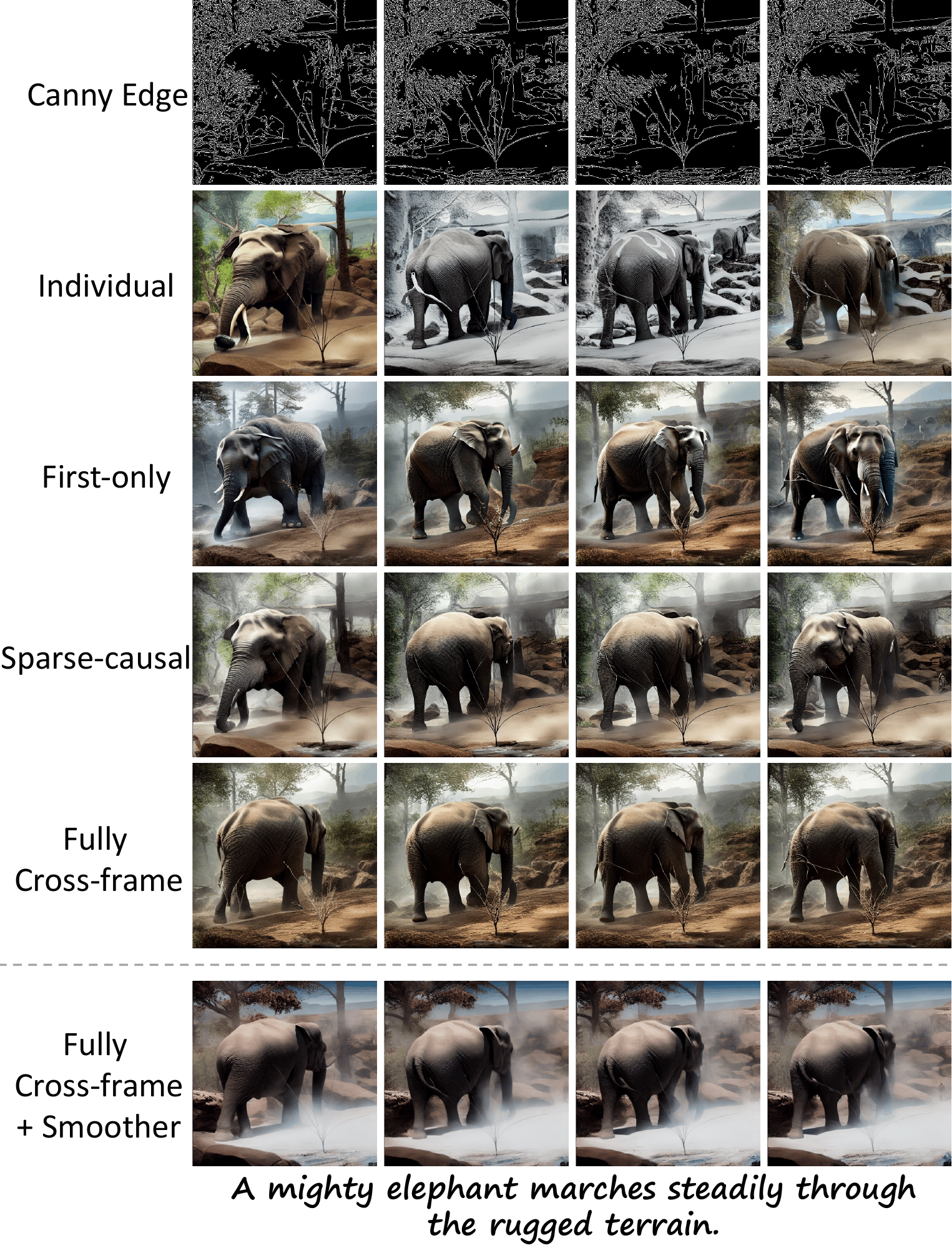}
    \caption{\textbf{Qualitative ablation studies} on cross-frame mechanisms and interleaved-frame smoother.
    Given canny edges in the first row, our fully cross-frame interaction produces video frames with higher quality and consistency than other mechanisms, and adding our smoother further enhances the video smoothness.
    }
    \label{fig:ab_crossframe}
\end{minipage}
\vspace{-1em}
\end{figure}

Fig.~\ref{fig:main_pose} further shows the comparison conditioned on human poses.
From Fig.~\ref{fig:main_pose}, Tune-A-Video only maintains the coarse structures of the source video, \ie, \texttt{human position}.
Text2Video-Zero and Follow-Your-Pose produce video frames with inconsistent appearance, \eg, 
\texttt{changing faces of iron man} (in row 4) or \texttt{disappearing objects in the background} (in row 5).
In comparison, our ControlVideo performs more consistent video generation, demonstrating its superiority.
More qualitative comparisons are provided in the supplementary materials.

\myparagraph{Quantitative results.}
We have also compared our ControlVideo with existing methods quantitatively on $125$ video-prompt pairs.
From Table~\ref{tab:main_consistency}, our ControlVideo conditioned on depth outperforms the state-of-the-art methods in terms of frame consistency and prompt consistency, which is consistent with the qualitative results.
In contrast, despite finetuning on a source video, Tune-A-Video still struggles to produce temporally coherent videos.
Although conditioned on the same structure information, Text2Video-Zero obtains worse frame consistency than our ControlVideo.
For each method, the depth-conditioned models generate videos with higher temporal consistency and text fidelity than the canny-condition counterpart, since depth maps provide smoother motion information.

\subsection{User study}
\begin{table}[t]
\caption{
    \textbf{User preference study.}
    The numbers denote the percentage of raters who favor the videos synthesized by our ControlVideo over other methods.}
    \vspace{2mm}
    \centering
    \scalebox{0.9}{
    \begin{tabular}{lccc}
    \toprule
         \makecell{Method Comparison} & \makecell{Video  Quality} & \makecell{Temporal Consistency} & \makecell{Text Alignment} \\
         % \Xhline{\arrayrulewidth}
         \midrule
         \makecell{Ours vs. Tune-A-Video~\cite{tune-a-video}} & $73.6\%$ & $83.2\%$ & $68.0\%$ \\
         % \Xhline{\arrayrulewidth}
         \makecell{Ours vs. Text2Video-Zero~\cite{khachatryan2023text2video}} & $76.0\%$ & $81.6\%$ &$65.6\%$\\
    \bottomrule
    \end{tabular}
    }
    % \vspace{-2mm}
    \label{tab:main_user}
\end{table}
We then perform the user study to compare our ControlVideo conditioned on depth maps with other competing methods. 
% \ie, ControlVideo (depth) vs. Tune-A-Video and ControlVideo (depth) vs. Text2Video-Zero (depth).
%
In specific, we provide each rater a structure sequence, a text prompt, and synthesized videos from two different methods (in random order).
Then we ask them to select the better synthesized videos for each of three measurements: (i) video quality, (ii) temporal consistency throughout all frames, and (iii) text alignment between prompts and synthesized videos.
The evaluation set consists of 125 representative structure-prompt pairs.
Each pair is evaluated by 5 raters, and we take a majority vote for the final result.
From Table~\ref{tab:main_user}, the raters strongly favor our synthesized videos from all three perspectives, especially in temporal consistency.
On the other hand, Tune-A-Video fails to generate consistent and high-quality videos with only DDIM inversion for structural guidance, and Text2Video-Zero also produces videos with lower quality and coherency.

\begin{table}[t]
   \centering
   \caption{\textbf{Quantitative ablation studies} on cross-frame mechanisms and interleaved-frame smoother.
    The results indicate that our fully cross-frame mechanism achieves better frame consistency than other mechanisms, and the interleaved-frame smoother significantly improves the frame consistency.
    }
   \resizebox{0.95\linewidth}{!}{
   \begin{tabular}{cccc}
      \toprule
    %   \cline{4-11}
      \makecell{Cross-Frame Mechanism} &
      \makecell{Frame Consistency ($\%$)} & \makecell{Prompt  Consistency ($\%$)} &\makecell{Time Cost (min)}\\
     \Xhline{\arrayrulewidth}
     % \midrule
      Individual &89.94 &30.79  &{1.2}\\
      First-only &94.92 &30.54  &{1.2}\\
      Sparse-Causal &95.06 &30.59 &{1.5}\\
      Fully &95.36 &30.76 &{3.0}\\
     \Xhline{\arrayrulewidth}
     % \midrule
      Fully + Smoother
      &\textbf{96.83} &\textbf{30.79}  &{3.5}	\\
      \bottomrule
   \end{tabular}
   }
    \label{tab:ab_crossframe}
    % \vspace{-1em}
 \end{table}
\subsection{Ablation study}
%
% We conduct ablation studies on different cross-frame mechanisms and the effectiveness of interleaved-frame smoother.
%
% We conduct ablation studies to evaluate the effects of various components in our ControlVideo.

\myparagraph{Effect of fully cross-frame interaction.} 
To demonstrate the effectiveness of the fully cross-frame interaction, we conduct a comparison with the following variants:
i) individual: no interaction between all frames,
ii) first-only: all frames attend to the first one,
iii) sparse-causal: each frame attends to the first and former frames,
iv) fully: our fully cross-frame, refer to Sec.~\ref{sec:method}.
%
% simply replacing self-attention modules with their cross-frame attention brings non-negligible generation gap, thereby decreasing the quality of synthesized frames.
%
Note that, all the above models are extended from ControlNet without any finetuning.
The qualitative and quantitative results are shown in Fig.~\ref{fig:ab_crossframe} and Table~\ref{tab:ab_crossframe}, respectively.
From Fig.~\ref{fig:ab_crossframe}, the individual cross-frame mechanism suffers from severe temporal inconsistency, \eg, \texttt{colorful and black-and-white frames}.
The first-only and sparse-causal mechanisms reduce some appearance inconsistency by adding cross-frame interaction.
However, they still produce videos with structural inconsistency and visible artifacts, \eg, \texttt{the orientation of the elephant and duplicate nose} (row 3 in Fig.~\ref{fig:ab_crossframe}).
In contrast, due to less generation gap with ControlNet, our fully cross-frame interaction performs better appearance coherency and video quality.
% attention concatenates all frames to become a ``larger image'', hence directly inherits high-quality and consistent generation from ControlNet.
%
% Considering the superior performance of ControlVideo, extra $1\sim 2 \times$ time cost is also acceptable.
%
Though the introduced interaction brings an extra $1\sim 2 \times$ time cost, it is acceptable for a high-quality video generation.

\myparagraph{Effect of interleaved-frame smoother.}
We further analyze the effect of the proposed interleaved-frame smoother.
From Fig.~\ref{fig:ab_crossframe} and Table~\ref{tab:ab_crossframe}, our interleaved-frame smoother greatly mitigates structural flickers and improves the video smoothness.

\subsection{Extension to long-video generation}
\begin{figure}[t]
   % \vspace{-1em}
  \begin{center}
  \includegraphics[width=.99\linewidth]{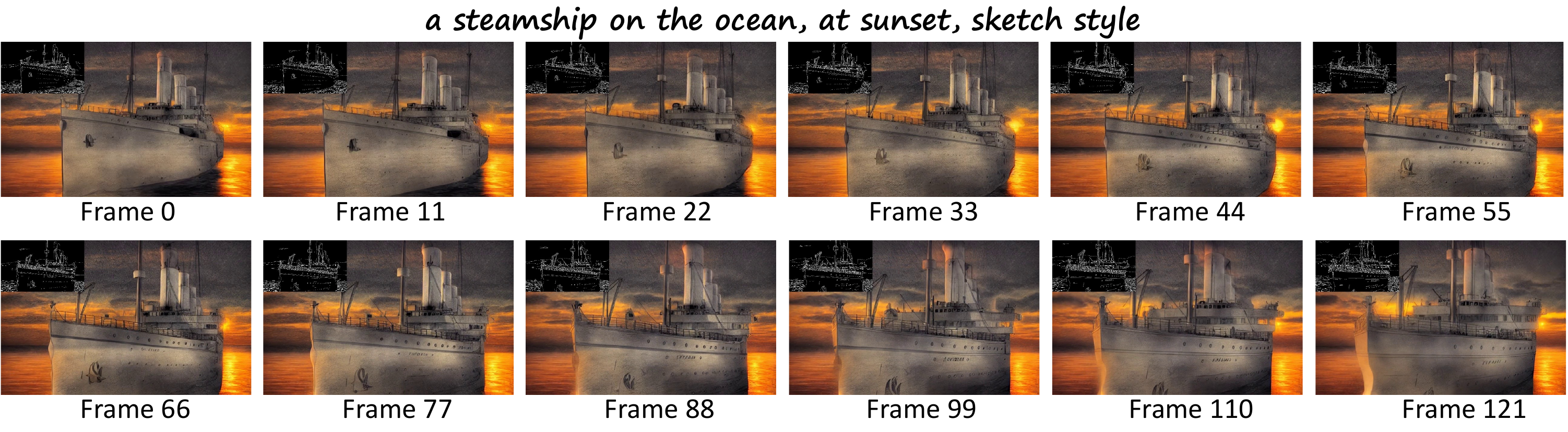}
  \end{center}
  \vspace{-2mm}
  \caption{
\textbf{A long video produced with our hierarchical sampling.}
Motion sequences are shown on the top left.
Using the efficient sampler, our ControlVideo generates a high-quality long video with the holistic consistency.
  \textbf{Results best seen at 500\% zoom.}
  }
    \label{fig:main_long}
  \vspace{-2.5mm}
\end{figure}
Producing a long video usually requires an advanced GPU with high memory.
%
% To enable long-video generation with commodity hardware, we propose a hierarchical sampling strategy to synthesize long videos in a memory-efficient manner.
With the proposed hierarchical sampler, our ControlVideo achieves long video generation (more than 100 frames) in a memory-efficient manner.
%
%  coupled with hierarchical sampling, 
As shown in Fig.~\ref{fig:main_long}, our ControlVideo can produce a long video with consistently high quality.
Notably, benefiting from our efficient sampling, it only takes approximately ten minutes to generate $100$ frames with resolution $512\times 512$ in one NVIDIA RTX 2080Ti.
More visualizations of long videos can be found in the supplementary materials.

\section{Related work}
\label{sec:related}
\myparagraph{Text-to-image synthesis.}
Through pre-training on billions of image-text pairs, large-scale generative models~\cite{nichol2021glide,balaji2022ediffi,saharia2022photorealistic,ramesh2022hierarchical,rombach2022high,ramesh2021zero,chang2023muse,ding2021cogview,ding2022cogview2,yu2022scaling,sauer2023stylegan,kang2023gigagan} have made remarkable progress in creative and photo-realistic image generation.
% 
% These generative models often follow the formulation of GANs~\cite{goodfellow2020generative,sauer2023stylegan,kang2023gigagan}, autoregressive models~\cite{nichol2021glide,chang2023muse,ding2021cogview,ding2022cogview2,yu2022scaling}, or diffusion models~\cite{DDPM_paper,balaji2022ediffi,saharia2022photorealistic,ramesh2022hierarchical,rombach2022high}.
%
Various frameworks have been explored to enhance image quality, including GANs~\cite{goodfellow2020generative,sauer2023stylegan,kang2023gigagan}, autoregressive models~\cite{nichol2021glide,chang2023muse,ding2021cogview,ding2022cogview2,yu2022scaling}, and diffusion models~\cite{DDPM_paper,balaji2022ediffi,saharia2022photorealistic,ramesh2022hierarchical,rombach2022high}.
%
% Among these generative models, diffusion-based models are well open-sourced and popularly applied to several downstream tasks, such as image editing~\cite{hertz2022prompt,meng2021sdedit}, customized generation~\cite{gal2022image,wei2023elite,kumari2022multi,ruiz2022dreambooth}, and controllable text-to-image generation~\cite{controlnet,mou2023t2i}.
%
% Controllable text-to-image models provide additional input conditions to pre-trained text-to-image diffusion models, so that they could produce images conditioned on both prompts and structures.
%
Among these generative models, diffusion-based models are well open-sourced and popularly applied to several downstream tasks, such as image editing~\cite{hertz2022prompt,meng2021sdedit} and customized generation~\cite{gal2022image,wei2023elite,kumari2022multi,ruiz2022dreambooth}.
Besides text prompts, several works~\cite{controlnet,mou2023t2i} also introduce additional structure conditions to pre-trained text-to-image diffusion models for controllable text-to-image generation.
Our ControlVideo is implemented based on the controllable text-to-image models to inherit their ability of high-quality and consistent generation.

\myparagraph{Text-to-video synthesis.}
Large text-to-video generative models usually extend text-to-image models by adding temporal consistency.
Earlier works~\cite{wu2022nuwa,hong2022cogvideo,wu2021godiva,villegas2022phenaki} adopt an autoregressive framework to synthesize videos according to given descriptions.
%
% With the unprecedentedly generative capabilities of diffusion models, recent works~\cite{imagen_video,video-diffusion-models,make-a-video} propose to 
% leverage them to produce high definition videos.
%
Capitalizing on the success of diffusion models in image generation, recent works~\cite{imagen_video,video-diffusion-models,make-a-video} propose to leverage their potential to produce high-quality videos.
Nevertheless, training such large-scale video generative models requires extensive video-text pairs and computational resources.

To reduce the training burden, 
Gen-1~\cite{gen1_paper} and Follow-Your-Pose~\cite{ma2023follow} provide coarse temporal information (\eg, motion sequences) for video generation, yet are still costly for most researchers and users.
By replacing self-attention with the sparser cross-frame mechanisms, Tune-A-Video~\cite{tune-a-video} and Text2Video-Zero~\cite{khachatryan2023text2video} keep considerable consistency in appearance with little finetuning.
Our ControlVideo also adapts controllable text-to-image diffusion models without any training, but generates videos with better coherency in both structure and appearance.

\section{Discussion}
\label{sec:discussion}
In this paper, we present a training-free framework, namely ControlVideo, towards consistent and efficient controllable text-to-video generation.
Particularly, ControlVideo is inflated from ControlNet by adding fully cross-frame interaction to ensure appearance coherence without sacrificing video quality.
Besides, interleaved-frame smoother interpolates alternate frames at sequential timesteps to effectively reduce structural flickers.
%
% Furthermore, hierarchical sampler is integrated into ControlVideo to facilitate efficient long-video synthesis. 
%
With the further introduced hierarchical sampler and memory-efficient designs, our ControlVideo can generate both short and long videos in several minutes with commodity GPUs.
%
% Both quantitative and qualitative experiments are conducted on extensive motion-prompt pairs, and show that ControlVideo performs better than previous state-of-the-arts in terms of video quality and temporal consistency.
%
Quantitative and qualitative experiments on extensive motion-prompt pairs demonstrate that ControlVideo performs better than previous state-of-the-arts in terms of video quality and temporal consistency.
%
% With the memory-efficient designs, our ControlVideo can generate both short and long videos in several minutes with commodity GPUs.

\myparagraph{Limitations.}
While our ControlVideo enables consistent and high-quality video generation, it still struggles with producing videos beyond input motion sequences.
For example, given sequential poses of \texttt{Michael Jackson's moonwalk}, it is difficult to generate a vivid video according to text prompts like \texttt{Iron man runs on the street}.
%
% In the future, we hope that motion sequences will be adapted to new ones based on input text prompts, so that users could create more diverse videos with our ControlVideo.
%
In the future, we will explore how motion sequences can be adapted to new ones based on input text prompts, so that users can create more diverse videos with our ControlVideo.

\myparagraph{Broader impact.}
Large-scale diffusion models have made tremendous progress in text-to-video synthesis, yet these models are costly and unavailable to the public.
Our ControlVideo focuses on training-free controllable text-to-video generation, and takes an essential step in efficient video creation.
Concretely, ControlVideo could synthesize high-quality videos with commodity hardware, hence, being accessible to most researchers and users.
For example, artists may leverage our approach to create fascinating videos with less time.
Moreover, ControlVideo provides insights into the tasks involved in videoss, \eg, video rendering, video editing, and video-to-video translation.
On the flip side, albeit we do not intend to use our model for harmful purposes, it might be misused and bring some potential negative impacts, such as producing deceptive, harmful, or explicit videos.
Despite the above concerns, we believe that they could be well minimized with some steps.
For example, an NSFW filter can be employed to filter out unhealthy and violent content.
Also, we hope that the government could establish and improve relevant regulations to restrict the abuse of video creation.

% \newpage
{\small
\bibliographystyle{ieee_fullname}
\bibliography{bib}
}

% \clearpage
% \input{text/appendix}
\end{document}